# Generalising Cost-Optimal Particle Filtering

Andrew Warrington[1] and Neil Dhir[1]

*Abstract*— We present an instance of the optimal sensor scheduling problem with the additional relaxation that our observer makes active choices whether or not to observe and *how* to observe. We mask the nodes in a directed acyclic graph of the model that are observable, effectively optimising whether or not an observation should be made at each time step. The reason for this is simple: it is prudent to seek to reduce sensor costs, since resources (e.g. hardware, personnel and time) are finite. Consequently, rather than treating our plant as if it had infinite sensing resources, we seek to *jointly* maximise the utility of each perception. This reduces resource expenditure by explicitly minimising an observation-associated cost (e.g. battery use) while also facilitating the potential to yield better state estimates by virtue of being able to use more perceptions in noisy or unpredictable regions of state-space (e.g. a busy traffic junction). We present a general formalisation and notation of this problem, capable of encompassing much of the prior art. To illustrate our formulation, we pose and solve two example problems in this domain. Finally we suggest active areas of research to improve and further generalise this approach.

## I. INTRODUCTION

Information retrieval is rarely free. Obtaining data may require requesting time on a radio telescope, performing invasive surgery or dropping a sensor beacon out at sea. Therefore it is intuitively reasonable to wish to minimise expenditure (running costs, resources, time etc.) while maximising the quality of the resulting state estimates. Therefore, in the most naive setting, we should aim to optimise the expected utility of each observation over influencing factors such as *when* and *what* we observe. This corresponds to optimising over which nodes of a graphical model we observe, as shown in fig. 1, where each observation improves our estimate of the latent state, but in so doing, incurs a finite cost. Consequently, active observers need to take measurements when it is prudent to do so, rather than greedily sampling continuously as happens in many domains.

This task is complicated by the fact that the utilities of any two observations are not independent from one another. In the scenario where we are able to make multiple observations we must consider the utility of *all* observations *jointly*. The solution of sequentially placing samples, each one optimal individually, yields a different solution from considering *all* observations jointly. Therefore, greedy optimisation of when to sample next is not a satisfactory solution in the general setting. The domain of this problem is typically filed under the *optimal scheduling problem*, which can also be viewed as a stochastic optimal control problem [1], [2], [3], [4].

[1]Department of Engineering Science, University of Oxford, England.
Correspondance: `andreww@robots.ox.ac.uk`

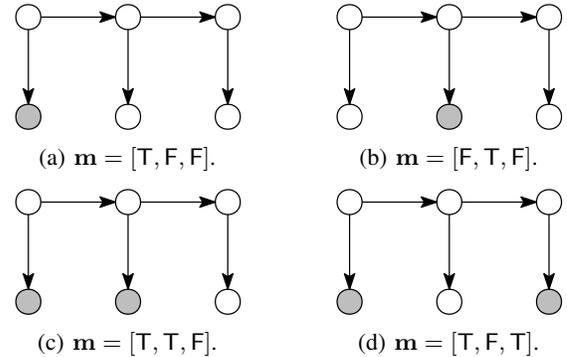

(a) $\mathbf{m} = [\mathsf{T}, \mathsf{F}, \mathsf{F}]$.    (b) $\mathbf{m} = [\mathsf{F}, \mathsf{T}, \mathsf{F}]$.

(c) $\mathbf{m} = [\mathsf{T}, \mathsf{T}, \mathsf{F}]$.    (d) $\mathbf{m} = [\mathsf{T}, \mathsf{F}, \mathsf{T}]$.

Fig. 1: Examples of state-space models with different observation masks, $\mathbf{m}$, where observed nodes are shaded and unobserved (latent) nodes are unfilled. Optimisation over $\mathbf{m}$, such that benefit is maximised while cost is minimised is the core topic of this paper.

### A. Problem definition

We seek to solve a maximisation of the following form:

$$\mathbf{m}^* = \arg\max_{\mathbf{m} \in \mathcal{M}} \{\mathcal{B}(\mathbf{m}) - \lambda \mathcal{C}(\mathbf{m})\}, \qquad (1)$$

where $\mathbf{m} \triangleq \{m_t \mid t = 1, \ldots, T\} \in \mathcal{M} = \{\mathsf{F}, \mathsf{T}\}^{1 \times T}$ corresponds to what we refer to herein as an *observation mask*, where $m_t = \mathsf{T}$ indicates that the $t^{\text{th}}$ node is observed, and $m_t = \mathsf{F}$ indicates it is not observed. The constant $T \in \mathcal{T} = \mathbb{Z}_+$ corresponds to the number of nodes in a graphical model which we can chose to observe or not, which directly influences the observation structure of the graphical model (see fig. 2). For a finite duration state-space model, $T$ corresponds to the number of time steps (assuming we are able to observe at every time step) yielding a one-to-one correspondence between mask elements and observable nodes as shown in fig. 2. The hyperparameter $\lambda \in \mathbb{R}_{\geq 0}$ allows one to upweight or downweight the importance of cost saving relative to good state estimation. For completeness, we do not preclude expansion of the domain of $m_t$ to allow for parameters required to make each observation to be jointly optimised alongside the mask itself.

The function $\mathcal{B} : \mathcal{M} \to \mathbb{R}$ represents some notion of the benefit of an observation mask. This may correspond to something as simple as the expected $L_2$ norm between the estimation of the state and the true state or, for example, the entropy [5] of the posterior distribution over the latent state. Naturally this depends on the observations that are made, i.e. *generally*, more observations lead to a better state estimate.

The function $\mathcal{C} : \mathcal{M} \to \mathbb{R}$ maps a particular mask onto a cost. This cost function may be as simple as the number of observations made, represented as $\Sigma_{t=1}^{T} \mathbb{I}[m_t \neq \mathsf{F}]$, where $\mathbb{I}(\cdot)$ is the indicator function. More complex variants may allocate more cost to observing certain nodes, having a cost that is dependent on state or even dependent on other elements in the mask itself. The benefit of a mask minus the associated cost is denoted as the *utility*:

$$\mathcal{U}(\mathbf{m}) = \mathcal{B}(\mathbf{m}) - \mathcal{C}(\mathbf{m}). \quad (2)$$

The main contribution of this work is casting the optimal sensor scheduling problem in this general framework, facilitating optimisation over the *number* of observations that can be made through the use of a mask; where our framework also facilitates the optimisation of parameters associated with making each observation. Although, solutions to similar problems have been presented previously [1], [6], [7], [8], [9], [10], [11], [12] there are some important differences between these studies and the techniques we discuss herein.

## II. PRIOR ART

Much of the prior art falls into one of two categories. First, scenarios where at each time step a measurement is taken using exactly one of a finite number of measurement beacons, each with different properties [1], [6], [7], [8], [9], referred to as *pool optimisations*. Second, other studies consider how best to manoeuvre a single observer over time such that the most accurate estimate of state is obtained [10], [11], [12], is referred to as *action optimisations*. Our framework generalises and unifies both pool optimisation and action optimisation approaches.

Further, many authors describe *when* and *where* to sample in a signal processing context as the aforementioned *sensor scheduling problem* [2], [7], [13]. This domain is gaining renewed interest as more and more autonomous systems come online [14], all of which have temporal and spatial *adaptive* sampling capability.

While there is a rich literature in each of these domains, it is the intersection of all domains that we target. To our knowledge, the thematically closest previous work is presented by [14]. In their study the authors address reducing the cost (in terms of energy consumption) of localising a mobile robot. This is achieved by scheduling localisations only when they are strictly necessary. In this case, 'strictly necessary' is defined as when the probability that the robot has left a pre-defined 'corridor' (centred on its intended trajectory) exceeds a threshold value. When this threshold is breached, an observation is made. This constraint is analogous to setting a maximum value, $W$, on the variance of the state estimate and rejecting any observation schedule that does not bound the variance to be within this value. Casting this constraint in our notation yields:

$$\mathcal{B}(\mathbf{m}) = \begin{cases} 0, & \text{if } \max \mathbb{V}[\mathbf{x}_t] < W, \ \forall t \in \{1, \dots, T\} \\ -\infty, & \text{otherwise,} \end{cases} \quad (3)$$

$$\mathcal{C}(\mathbf{m}) = \Sigma_{t=1}^{T} \mathbb{I}[m_t \neq \mathsf{F}], \quad (4)$$

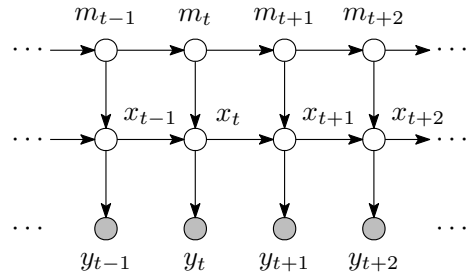

Fig. 2: A standard state-space model augmented with mask parameters noted, in our notation, by $\{m_t \mid t = 1, \dots, T\}$. The mask parameters dictate *if* an observation should be made ($m_t \neq \mathsf{F}$).

where $\mathbb{V}[\mathbf{x}_t]$ is the variance of the state estimate at time $t$. While the method by [14] is a novel solution, there are several notable drawbacks with the suggested algorithm.

The main drawback is that the construct of a corridor of acceptability is unnatural, especially when applied to other problem domains. A more natural metric, especially for tracking, and one that has been used previously in sensor scheduling, is minimising the squared reconstruction error over the whole trajectory. The use of such a metric is not supported under [14].

*An aside on particle filters:* Throughout this work we make use of a particle filter to provide state estimates conditioned on the observed data. Exhaustive exegesis of particle filtering is outside the scope of this paper. We refer the reader to [15] and §27 of [16] for more extensive details.

## III. GRAPHICAL MODEL REPRESENTATION

Typically the sensor scheduling problem is exposed as a hidden Markov model (HMM) [10], [2], [17], [18]. Though there are other ways of representing the problem, probabilistic graphical models are a useful foundation as they model the conditional dependence structure between random variables. Since we have a number of random variables, for which we seek a posterior estimate, this construction is thus preferred.

We present the model construction in fig. 2 which is similar to the switching state-space model (SSSM) [19], which has previously been used for target tracking [2]. Although the SSSM is a more complex version of the standard state-space model, the only addition is a discrete switch variable that indexes the mode. The latent states and the observations (measurements) are denoted by $x_t$ and $y_t$ respectively. Similar to the SSSM we add an additional latent layer (top layer in fig. 2) to the standard SSM, but instead of modelling the switching dynamics, we model the evolution of the binary mask $m_t$ at each time-step. The mask is only active ($m_t \neq \mathsf{F}$) when the utility warrants it.

## IV. METHOD

We now introduce two synthetic examples, both in the context of cost-optimal target tracking in a HMM.

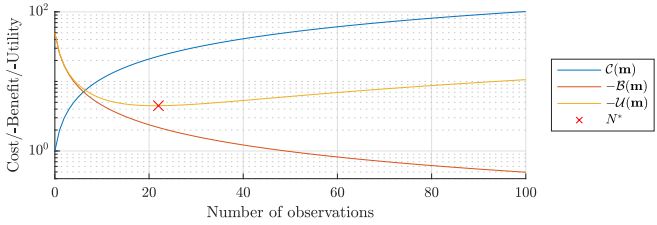

(a) The components of eq. (1) and their sum. Here, we plot the negative of eq. (1) and minimise the negative utility due to the polarity of the definitions in eq. (5) and eq. (6). The optimal number ($N^* = 22$) is indicated by a cross ($\times$).

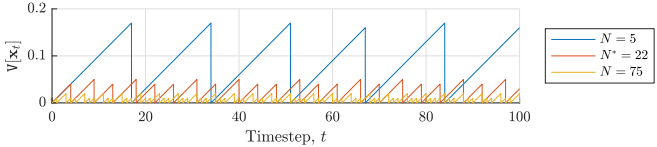

(b) The evolution of variance with different numbers of samples. The blue line indicates the evolution when too few samples are taken ($N = 5$), incurring little cost, but also poorly estimating state. The orange line indicates when too many samples are taken ($N = 75$), well estimating state, but incurring high cost. The red line shows the evolution when the optimal number of samples are taken ($N^* = 22$), balancing cost saving and accuracy requirements.

Fig. 3: Cost-benefit analysis of the regular sampling problem.

### A. Regular sampling problem

Suppose we are tracking an object for a duration $T$, and that making an observation has a fixed cost $C_0$. Accordingly, we wish to optimise the number of observations we make, $N$, such that we estimate the state well, while not incurring unnecessary cost. Casting this in our formalism:

$$\mathcal{C}(\mathbf{m}) = C_0 \Sigma_{t=1}^T \mathbb{I}[m_t \neq \mathsf{F}], \tag{5}$$
$$\mathcal{B}(\mathbf{m}) = -\Sigma_{t=1}^T \mathbb{V}[\mathbf{x}_t], \tag{6}$$

where the function $\mathbb{V}[\mathbf{x}_t]$ calculates the variance of all the particles at time $t$.

For simplicity, we consider a model with a Gaussian random walk on velocity, with variance $\sigma_p^2$, noise-free observations and perfect knowledge of the initial state, tantamount to the model used in [14]. The variance is analytically computable (in expectation) as $\mathbb{V}[\mathbf{x}_t] = \sigma_p^2(t - \tau)$, where $\tau$ is the time stamp of the most recent observation. We set $\tau$ to zero if no observation has been made yet.

By enumerating over the number of samples that can be made, we can solve for the optimum number of samples to make. This can be seen in fig. 3a. The blue line indicates the cost of the mask, $\mathcal{C}(\mathbf{m})$, equal the number of observations that are made, while the red line indicates the benefit of the mask, $\mathcal{B}(\mathbf{m})$, equal to the sum of the variances over the course of the trajectory. The sum of these terms is displayed in orange, where the optimum is indicated with a red cross. The evolution of the variance of the state estimate with time is displayed in fig. 3b for different values of $N$. Intuition about the problem can be gained by noting that the area under each trace is the value of $-\mathcal{B}(\mathbf{m})$, while the number of samples is the cost $\mathcal{C}(\mathbf{m})$.

### B. One Sample Problem

Consider the case of localising a submarine travelling along a known trajectory, but with an unknown velocity profile, for fixed duration $T$. We wish to localise the vessel, such that the reconstruction error is low, but constrain ourselves to only being able to observe *once*, at time $t_b$. Further, we introduce the notion (akin to helicopter dipping sonar systems) that we can *place* the sensor at $x_b$, where the placement influences the noise model. To put it another way, we wish to estimate *a priori* the best time and position to localise at, given we have just one observation, over all possible trajectories. Therefore, we must jointly optimise over $t_b$ and $x_b$.

As the submarine is progressing along a fixed trajectory, we consider this a 1D problem, tracking how far along the trajectory the vessel is at any point in time, where the submarines' velocity profile evolves according to a Gaussian random walk. This assumption is directly analogous to the assumption used in [14]. We denote the true position of the submarine at time $t$ as $x(t)$, and the overall profile is denoted as $\mathbf{x} = [x(1), x(2), \dots, x(t), \dots, x(T)]$. We denote any estimate as a subscript $x_t$, particles are indicated by a superscript index $\mathbf{x}^{(j)}$ and expectations by a hat $\hat{\mathbf{x}}$.

We now formalise this example in the nomenclature introduced in §I. The mask $\mathbf{m}$, is a $T$ length vector, where the $t_b^{\text{th}}$ entry is not equal to $\mathsf{F}$ but is instead equal to $x_b$. As an example: $\mathbf{m} = [\mathsf{F}, \mathsf{F}, \mathsf{F}, 5, \mathsf{F}, \dots, \mathsf{F}]$, here $t_b = 4$ and $x_b = 5$. The corresponding cost and benefit functions are:

$$\mathcal{C}(\mathbf{m}) = \begin{cases} 0, & \text{if } \Sigma_{t=1}^T \mathbb{I}[m_t \neq \mathsf{F}] = 1 \\ \infty, & \text{otherwise.} \end{cases} \tag{7}$$
$$\mathcal{B}(\mathbf{m}) = -\mathbb{E}_{\mathbf{x}}\left[||\hat{\mathbf{x}} - \mathbf{x}||_2^2\right], \tag{8}$$

where $\mathbf{x}$ is the true trajectory.

The observation model, denoted $g(x(t), x_b) : \mathbb{R} \times \mathbb{R} \to \mathbb{R}$, generates a measurement $y \in \mathbb{R}$ is given by:

$$y \sim g(x(t_b), x_b) \triangleq \mathcal{N}(x(t_b), \sigma_m^2) \tag{9}$$
$$\sigma_m^2 \triangleq \alpha|x(t_b) - x_b| + \epsilon, \tag{10}$$

where $\alpha$ and $\epsilon$ are known constants.

We now discuss how we solve this problem. First, particles are initialised from the model and forward simulated under the model until the observation time $t_b$. An observation is then received and hence we wish to perform re-sampling. To perform the re-sampling step in the particle filter, we must know the variance associated with the observation, $\sigma_m^2$, given by eq. (10). However, we cannot evaluate this exactly as we do not know the position of the submarine. Therefore we resort to estimating it by marginalising over the true trajectory. Fortunately, we already have the required samples to perform this marginalisation, as the set of particles at $t_b$ are distributed according to the true distribution. Therefore, we calculate an estimate of the variance of the measurement

**Algorithm 1:** One Sample Problem

**Input :** Buoy deployment position $x_b$, buoy deployment time $t_b$, Monte Carlo samples $K$, particle count $J$.

$\widehat{B} \leftarrow 0$ ▷ *Initialise cost score.*
**for** $k \in \{1, \ldots, K\}$ **do**
    $\mathbf{x} \sim p(\mathbf{x})$ ▷ *Draw truth from model.*
    $y \leftarrow \mathtt{simulate\_observation}(x_b, x(t_b)))$
    $x_0^{(j)} \sim p(x_0), \ \forall j \in \mathcal{J}$ ▷ *Initialise particles.*
    **for** $t \in \{1, \ldots, t_b\}$ **do**
        $x_t^{(j)} \sim p(x_t^{(j)} \mid x_{t-1}^{(j)})$ ▷ *Iterate particle.*
    $\widehat{\sigma}_m^2 \leftarrow \mathbb{E}\left[\alpha|x_{t_b}^{(i)} - x_b| + \epsilon\right]$ ▷ *Expected variance.*
    $\mathbf{x}_t \leftarrow \mathtt{resample}(\mathbf{x}_t, y, \widehat{\sigma}_m^2)$
    **for** $t \in \{t_b + 1, \ldots, T\}$ **do**
        $x_t^{(i)} \sim p(x_t^{(j)} \mid x_{t-1}^{(i)})$ ▷ *Iterate particle.*
    $\widehat{\mathbf{x}} \leftarrow \mathbb{E}[\mathbf{x}^{(j)}]$ ▷ *Take expectation at each time.*
    $\widehat{B} \leftarrow \widehat{B} + \|\widehat{\mathbf{x}} - \mathbf{x}\|_2^2/K$

**Output:** $-\widehat{B}$

as:

$$\widehat{\sigma}_m^{2(j)} = \alpha|x_{t_b}^{(j)} - x_b| + \epsilon, \quad (11)$$

$$\widehat{\sigma}_m^2 = \mathbb{E}[\widehat{\sigma}_m^{2(j)}] \quad (12)$$

where the functional form of eq. (11) is identical to that of eq. (10). We then simulate an observation (using the true location), re-sample the particles and iterate for the remaining $T - t_b$ steps. We repeat the analysis with $K$ synthetic trajectories, thereby marginalising over the true trajectory.

The output, $-\widehat{B}$, from Algorithm 1 is a Monte Carlo estimate of the (negative) expected reconstruction error for a given $(t_b, x_b)$ pair. For simplicity, we perform a grid search to optimise over $x_t$ and $x_b$. The results of this optimisation are shown in fig. 4. The height and colour of the surface in fig. 4 indicates the expected reconstruction error, $-\mathcal{B}(\mathbf{m})$. The single global optimum exists at $(t_b^*, x_b^*) = (6, 11.91)$.

## V. DISCUSSION

A simple extension to the work is to generalise the one sample problem to $N$ samples. While grid search may be plausible for small $N$, it rapidly becomes infeasible as the search space grows and so more sophisticated optimisers such as simulated annealing or Bayesian optimisation should be used for larger search spaces.

Further, this estimation is referred to as a 'dead-reckoned' solution, as it is solved for upfront and there is no scope for information from observations to be utilised. This is important as the trajectory of the object may abruptly change, resulting in observations that are poorly explained under the current state, and hence more observations must be made to correct the state estimate (possibly at the expense of reduced accuracy later). This behaviour can be achieved by estimating a computationally cheaper 'perturbation' to the previous mask by repeating the above analysis using the dead-reckoned mask from the previous step as an initialisation. This extension would also allow $N$ to also be a variable we optimise over on-the-fly, i.e. 'cancelling' observations if our state estimate is better than expected, or adding observations if the state estimate is poor.

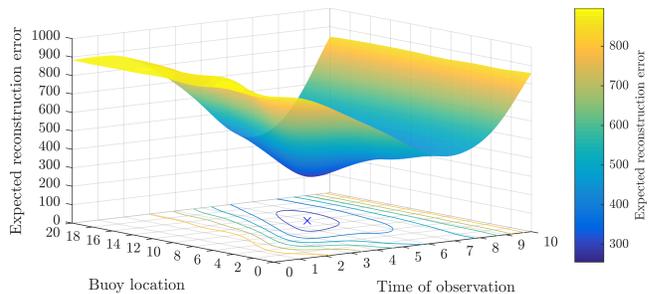

Fig. 4: Expected reconstruction error, $\mathcal{B}(\mathbf{m})$, of the one sample problem. The surface shown is a Gaussian process regression to create a smooth surface. The optimal parameter settings are indicated by a cross ($\times$).

However brute-force enumeration, Metropolis-based or simulation-based methods rapidly become computationally infeasible for large state-space models. Therefore, more efficient methods must be developed as the problem evolves. Such a method may be continuously relaxing the discrete distribution using the relaxation presented in [20] and [21], such that $\mathbf{m}$ instead contains values in the range $[0, 1]$ representing the weights in a weighted sum. These weights are then annealed to be a discrete distribution. This relaxation permits gradient steps to be taken in a discrete space which are often more informative than Metropolis steps.

## VI. CONCLUSION

We have presented a methodology for quantifying and controlling the accuracy of the target state estimate, such that the accuracy of the estimation is traded-off against the cost of the observation policy. The guiding paradigm in this work is that we are resource constrained, that we have a certain finite budget to which we must conform. All non-autonomous (input dependent) systems have finite control resources, consequently, it is necessary to consider the budget for these, such that cost-optimal system performance can follow.

Our approach ultimately has utility in any localisation task where we have control over our sensor infrastructure and making observations incurs some cost, or, is of variable benefit in different regions of statespace, such that the utility of *all* observations must be maximised. We have shown how two simple problems can be solved to both serve as illustration of the flexibility of our method, but also to explore how to solve more complex problems. We then suggest active areas of research for improving and generalising the approach further.